\documentclass[runningheads]{llncs}
\usepackage{graphicx}
\usepackage{hyperref}
\usepackage{booktabs}
\usepackage{lmodern}
\usepackage[T1]{fontenc}

\newcommand{\ie}{{\it i.e.}}
\newcommand{\eg}{{\it e.g.}}

\begin{document}
%
%\title{SIGMA: Human-Labeled Semantic Parsing and Statistical Analysis Dataset for Text-to-Code}
\title{SIGMA: A Dataset for Text-to-Code Semantic Parsing with Statistical Analysis}


\author{Saleh Almohaimeed*\ \and
Shenyang Liu* \and
May Alsofyani* \and
Saad Almohaimeed \and
Liqiang Wang}
\authorrunning{S. Almohaimeed et al.}
\titlerunning{SIGMA: Text-to-Code dataset}
\institute{Department of Computer Science, University of Central Florida, USA \\
\email{\{saleh, shenyang, may\_sof, saadm\}}@knights.ucf.edu,  \email{lwang}@cs.ucf.edu\\
}

\maketitle              % typeset the header of the contribution
\renewcommand{\thefootnote}{}
\footnotetext{\makebox[\columnwidth]{1946-0759/23/\$31.00~\copyright2023 IEEE \hfill} \hspace{\columnsep}\makebox[\columnwidth]{ }}

\begin{abstract}
In the domain of semantic parsing, significant progress has been achieved in Text-to-SQL and question-answering tasks, both of which focus on extracting information from data sources in their native formats. However, the inherent constraints of their formal meaning representations, such as SQL programming language or basic logical forms, hinder their ability to analyze data from various perspectives, such as conducting statistical analyses. To address this limitation and inspire research in this field, we design SIGMA, a new dataset for Text-to-Code semantic parsing with statistical analysis. SIGMA comprises 6000 questions with corresponding Python code labels, spanning across 160 databases. Half of the questions involve query types, which return information in its original format, while the remaining 50\% are statistical analysis questions, which perform statistical operations on the data. The Python code labels in our dataset cover 4 types of query types and 40 types of statistical analysis patterns. We evaluated the SIGMA dataset using three different baseline models: LGESQL, SmBoP, and SLSQL. The experimental results show that the LGESQL model with ELECTRA outperforms all other models, achieving 83.37\% structure accuracy. In terms of execution accuracy,  the SmBoP model, when combined with GraPPa and T5, reaches 76.38\%.

\def\thefootnote{*}\footnotetext{These authors contributed equally to this work.}\def\thefootnote{\arabic{footnote}}
\def\thefootnote{}\footnotetext{Our dataset avaliable at github.com/sasmohaimeed/SIGMA}\def\thefootnote{\arabic{footnote}}

\keywords{Semantic Parsing  \and Statistical Analysis \and Text-to-Code}
\end{abstract}
\section{Introduction}
An important field of natural language processing is semantic parsing (SP), which focuses on translating natural language sentences into machine-interpretable meaning representations. 
These target representations can be various forms such as specific programming language (SQL), formal mathematical expression (lambda calculus), or simple representation (text). The focus of most SP tasks is on retrieving information from knowledge sources \eg, databases, web pages, without much consideration for the use of statistical analysis to explore the data in different ways. The Text-to-SQL task is one of the most popular tasks in semantic parsing with few statistical capabilities. It is limited to three statistical functions (Sum, Average, and Count), making it incapable of performing more advanced statistical calculations. 

In order to accommodate diverse data exploration requirements and emphasize the most commonly used statistical functions, we design SIGMA, a cross-domain dataset with 160 databases and  6,000 natural language questions, each paired with corresponding Python code labels. SIGMA encompasses a total of 40 statistical analysis patterns, enabling the execution of statistical functions on data prior to presenting the results to users. Furthermore, there are 4 types of query patterns that resemble SQL language clauses, which directly retrieve data from the database. Within SIGMA's 6,000 questions, 3,000 statistical questions were written by nine individuals who are either in their final year of undergraduate studies or holding a degree in statistics. Within the 3,000 query questions, 2000 of these were composed by three graduate computer science students, and 1000 questions were taken from the Spider \cite{yu2018spider} dataset to ensure a variety range of questions. Given Python's ability to conduct various types of statistical analysis and perform operations beyond that scope, we chose it as the target language. %There are five Python parameters associated with each question that can be executed using our built-in Python executer.

To evaluate the quality and diversity of this dataset, we conducted experiments using three semantic parsing models: LGESQL \cite{cao2021lgesql}, SLSQL \cite{lei2020re}, and SmBoP \cite{rubin2020smbop}. Our experimental findings indicate that the LGESQL model delivers the highest performance in structural accuracy, while the SmBoP model attains good results in execution accuracy.

This paper makes the following contributions. 
(1) We design SIGMA, a dataset that consists of 6,000 questions over 160 databases. In total, the dataset covers 44 distinct patterns, including 40 statistical analysis patterns and 4 kinds of SQL clauses.
(2) We develop a built-in Python executor capable of executing all 44 patterns featured in our dataset.
(3) Experiments are conducted on the dataset using three models: LGESQL \cite{cao2021lgesql}, SLSQL \cite{lei2020re} and SmBoP \cite{rubin2020smbop}. 

\vspace{-0.4cm}
\section{Related Work}
\vspace{-0.2cm}

Over the past two decades, many semantic parsing and code generation tasks have been introduced to address specific user needs. The target meaning representations (MRs) vary from one task to another. While some MRs are simple, such as sentences or numbers, others are more complex, such as programming languages or multiple paragraphs. Our work, text-to-code, is relevant to three semantic parsing tasks: Text-to-SQL, question-answering, and code generation tasks.

For the Text-To-SQL task, an extensive amount of research has been conducted on the process of querying and retrieving information from relational databases. The Text-to-SQL task attempts to map natural language questions into executable SQL queries. Early research on this task was based on small and single-domain datasets such as ATIS \cite{price1990evaluation} \cite{dahl1994expanding}, Academic \cite{li2014constructing}, and Scholar \cite{iyer2017learning}. Followed by studies on large datasets such as WikiSQL \cite{zhong2017seq2sql} and more complex cross-domain datasets like Spider \cite{yu2018spider}. Recent research continues to address the challenge of handling cross-domain datasets as well as dealing with multiple interrelated natural languages queries such as SParC \cite{yu2019sparc} and CoSQL \cite{yu2019cosql}.

The purpose of Question-Answering task is to answer questions from a given context. The context can be one single document like in SQuAD \cite{rajpurkar2018know} or multiple documents like in TriviaQA \cite{joshi2017triviaqa}. The context can be stored in a structured or semi-structured knowledge base like in QAngaroo \cite{welbl2018constructing} and WebQuestions \cite{berant2013semantic} datasets. To accomplish this task, models with advanced reasoning capabilities should be devised to comprehend the given context and answer the user's questions.

For the code generation task, there are a wide range of subtasks within the code generation task, including repairing code, translating from code to code, completing code, and generating code from text. We focus on text-to-code generation in this research, in which a text sequence is translated into a code sequence. A few datasets were created in this task, including Card2code \cite{ling2016latent} that maps descriptions of cards to codes that implement them using the Python programming language. The disadvantage of Card2code is that it is too domain-specific. This domain-specific limitation is addressed by the Django \cite{oda2015learning} dataset. It contains the entire Django web framework source code with English annotations for each line of code. In the CoNaLa \cite{yin2018learning} dataset, the natural language intents of developers are mapped to code snippets collected from Stack Overflow. To enhance model performance on this task, researchers should explore strategies for establishing parallel alignments between natural language queries and the corresponding code. This can be achieved by implementing a set of rules to constrain the generation of code.

In semantic parsing, information is extracted from a variety of sources, such as databases, knowledge graphs, documents, and web pages. However, current approaches do not leverage the features of current programming languages and manipulate the data in the user's favor before it is retrieved. One of our motivations for this paper is to address the need by developing a dataset with an executor that can analyze the data statistically and explore it from a variety of perspectives.

\vspace{-0.2cm}
\section{Dataset}
\vspace{-0.2cm}

There are 6,000 questions in the SIGMA dataset, each with corresponding Python code as the ground truth. Nine people with degrees in statistics or related fields wrote 3,000 statistical questions. Of the remaining 3000 questions, 2000 were written by three graduate students in computer science, and 1000 questions were taken from Spider \cite{yu2018spider} dataset. 

The statistical analysis field contains a large number of different patterns. We conducted extensive research to determine what patterns to include. \cite{bruce2020practical} provides a detailed explanation of how to apply different statistical patterns to data science. From \cite{bruce2020practical}, we chose all applicable statistical analysis patterns for our dataset and classify them into three categories, \ie, distribution, plot, and numeric. Distribution includes functions that show curves indicating all possible value positions for a given data variable, such as normal, exponential, and chi-square distributions. Plot includes graphical techniques for representing one or more data variables such as histogram, hexbin, and contour techniques. The results of distribution and plot categories will be presented using different types of figures. The numerical category includes all other statistical calculations whose results can be expressed as tables or numbers, such as mean, correlation matrix, and frequency table. In addition to statistical analysis, we have considered different types of queries for databases, which include four SQL clauses: SELECT, WHERE, GROUP BY, and ORDER BY. \autoref{tab:results} shows all the statistical and SQL patterns included in our dataset.

% \usepackage{booktabs}
% \usepackage{graphicx}
\begin{table}[]
\centering
\caption{This table displays all the statistical and SQL patterns contained in our dataset. The type of each pattern refers to the Main-Kind type.}
\label{tab:patterns}
\resizebox{\textwidth}{!}{%
\begin{tabular}{@{}cc|cc|cc@{}}
\toprule
\textbf{Pattern}   & \textbf{Type}      & \textbf{Pattern}                            & \textbf{Type}             & \textbf{Pattern}   & \textbf{Type}      \\ \midrule
Select          & Query        & Box             & Plot    & Outlier                 & Numeric \\
Where           & Query        & KDE             & Plot    & Standard Deviation      & Numeric \\
Order by        & Query        & Pie             & Plot    & Variance                & Numeric \\
Group By        & Query        & Bar             & Plot    & Range                   & Numeric \\
Noraml          & Distribution & Scatter         & Plot    & Interquartile Range     & Numeric \\
Standard Normal & Distribution & Hexbin          & Plot    & Frequency Table         & Numeric \\
Long Tailed     & Distribution & Contour         & Plot    & Mode                    & Numeric \\
Binomial        & Distribution & Violin          & Plot    & Standard Error          & Numeric \\
Poisson         & Distribution & Mean            & Numeric & Percentile              & Numeric \\
Exponential     & Distribution & Weighted Mean   & Numeric & Correlation Matrix      & Numeric \\
Weibull         & Distribution & Trimmed Mean    & Numeric & Correlation Coefficient & Numeric \\
Chi-Square           & Distribution         & Mean Absolute Deviation                       & Numeric                     & Contingency table    & Numeric              \\
T               & Distribution & Median          & Numeric & Size                    & Numeric \\
F               & Distribution & Weighted Median & Numeric & Confidence Interval     & Numeric \\
Histogram            & Plot                 & \multicolumn{1}{l}{Median Absolute Deviation} & \multicolumn{1}{l}{Numeric} & \multicolumn{1}{l}{} & \multicolumn{1}{l}{} \\
\multicolumn{1}{l}{} & \multicolumn{1}{l}{} & \multicolumn{1}{l}{}                          & \multicolumn{1}{l}{}        & \multicolumn{1}{l}{} & \multicolumn{1}{l}{} \\ \bottomrule
\end{tabular}%
}
\end{table}

\vspace{-0.6cm}
\subsection{Components of Label (Python Code)}
\vspace{-0.2cm}

\begin{figure}[h]
    \centering
    \includegraphics[width=1\textwidth]{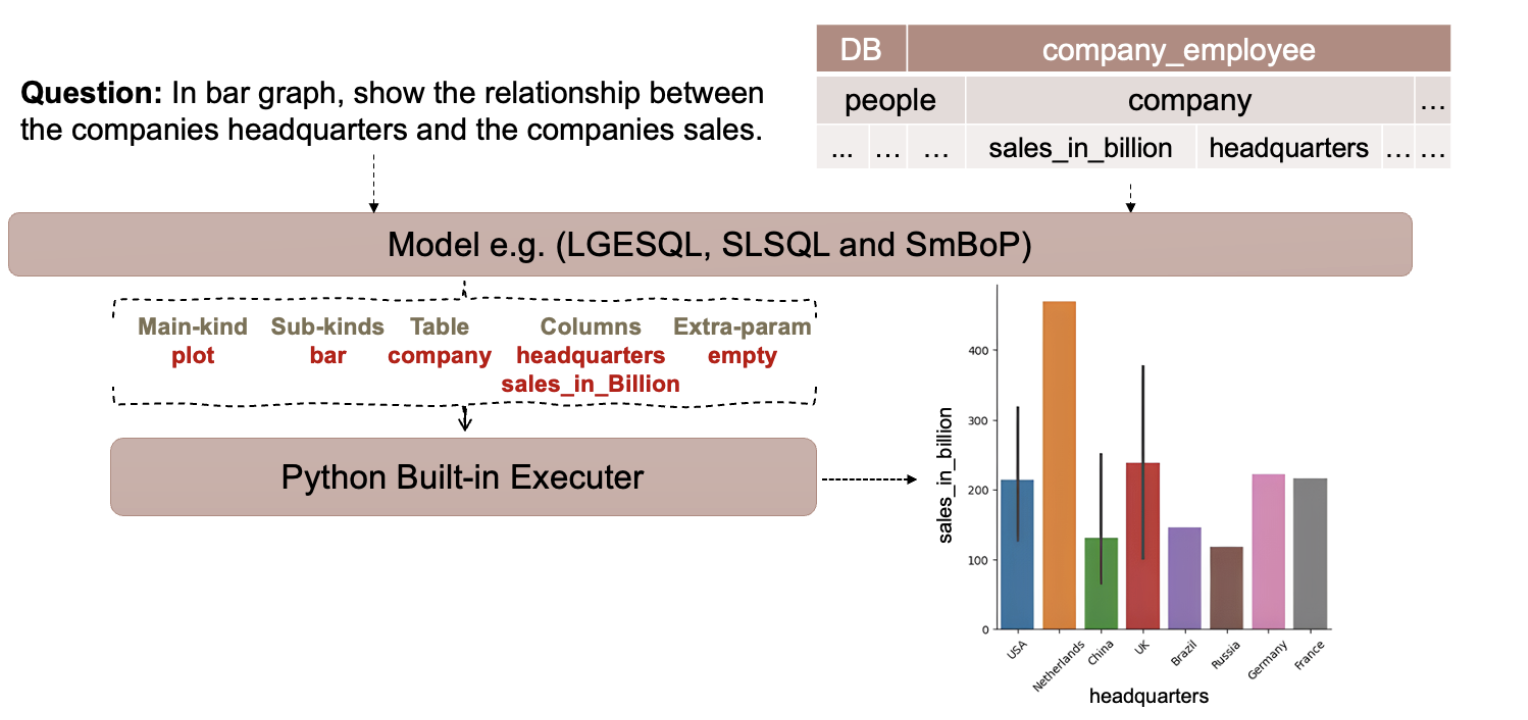}
    \caption{The overall architecture of our text-to-code task. The questions and the schema tables and columns are the inputs for the models, while the Python code labels are the outputs of the models. To verify the results, the  Python code can be executed using the built-in Python executer and results will be presented.}
    \label{fig:mesh2}
\end{figure}

Our goal is to map natural language text into a Python code snippet that will be inserted later into Python functions and executed. Python code snippet includes five components, as shown in \autoref{fig:mesh2}. The first component is {\it main-kind}, which can be one of four values: Distribution, Plot, Numeric, or Query. The second is known as {\it sub-kinds}, which vary depending on the main-kind. There are 10, 9, 21, and 4 different sub-kinds for Distribution, Plot, Numeric, and Query, respectively. The third component is related to the database {\it table}. In the fourth component, the selected {\it columns} from the database are indicated. {\it Extra-param} is the fifth component, which includes additional information required by the sub-kind. For example, when the sub-kind is labeled as ``orderby'', it is essential to specify whether the order is ascending or descending in the extra-param. However, most sub-kinds do not necessitate an extra-param, so they can be left empty.

\vspace{-0.2cm}
\subsection{Question and Python Code Construction}

%According to the type of questions, we have collected the questions and Python parameters differently. For the query questions, we asked three graduate students in computer science who are very knowledgeable about SQL programming to explore the databases and create questions with corresponding Python parameters. Questions for statistical analysis patterns were designed by nine individuals who are either in their final year of undergraduate studies or holding a degree in statistics. It was not the statistical people decision to choose the columns in the database, but rather it was provided to them in an excel sheet that contains the database name, table name, and columns names. Additionally, we have allowed them to explore the databases in order to see the values in the tables. Their job was only to write the questions. We have ensured that all individuals who wrote the questions followed the following rules.

When creating questions, the designers are not given the freedom to choose the columns in the database at their will. Instead, we provide them an excel sheet that contains the database name, table name, and column names. Additionally, we allow them to explore the databases in order to view the values in the tables. Their job is only to write the questions. We ensure that all individuals who wrote the questions follow the rules below.

\vspace{-0.4cm}
\subsubsection{Question Clarity.}
   An ambiguous question refers to a question in which different people interpret the same word or phrase differently. For example, if the question was ``Calculate the mean of the employee table'', it is impossible to calculate the mean of the employees if the column is unknown. A better question would be ``Calculate the mean of the employees' income''. In addition, there are no questions that require the use of outsources, such as "Calculate the mean salary of the hard workers". If the database does not have any information regarding the amount of work that each employee performs, then it is necessary to refer to outsource information that indicates which employees are hard workers. Hence, such questions with ambiguity are not included in our dataset.
   
   \vspace{-0.4cm}
   \subsubsection{Question Synonyms.}
   Each participant has been asked not to indicate the exact column, table, and statistical patterns names in each question. There should be at least 20 percent of questions that contain synonyms. As an example, if a column was named ``salary'', a participant should use other synonyms such as ``income'' or ``wage'' for some questions. Additionally, participants should be aware that they do not write all the questions in the same manner, such as mentioning the statistical patterns always before the columns or starting with ``what are'' phrase for every question.

\subsection{Dataset Statistics}
In Text-to-SQL tasks, the inputs to the models are the questions and schemas, which are the same as inputs to the models in Text-to-Code task. Therefore, in \autoref{tab:comparison} we summarize a comparison of our dataset with other popular Text-to-SQL datasets.

 Note that the number of questions in the SIGMA dataset is not as many as other datasets such as the Spider data \cite{yu2018spider}, but there is a great deal of diversity in the questions. Compared to Spider \cite{yu2018spider}, which was annotated by 11 students, our dataset contains 5000 questions annotated by 12 individuals who are experts in the field of SQL language or statistics. Additionally, 1000 questions were taken from the Spider dataset. As a result of this type of diversity, the models learn more and gain a greater understanding of the meaning of the questions. Furthermore, most other datasets have a single database within a single domain. SIGMA has 160 different databases across 107 different domains. Comparing to the Spider dataset, SIGMA does not duplicate any domain in the training and testing datasets. %Moreover, a total of 44 patterns are included in the SIGMA dataset. 
While the dataset is not designed for the Text-to-SQL task, it contains similar query types as most other Text-to-SQL datasets, including SELECT, WHERE, GROUPBY, and ORDERBY as well as 40 other statistical analysis patterns.

% \usepackage{booktabs}
% \usepackage{graphicx}
\vspace*{-10pt}
\begin{table}[h]
\centering
\caption{Comparisons of SIGMA dataset with popular text-to-SQL datasets.}
\label{tab:comparison}
\resizebox{\textwidth}{!}{%
\begin{tabular}{@{}ccccccc@{}}
\toprule
\textbf{Dataset}     & \textbf{\# Q}        & \textbf{\# labels}   & \textbf{\# DBs}      & \textbf{\# Domains}  & \textbf{\# patterns}  & \textbf{\# annotators} \\ \midrule
\multicolumn{1}{c|}{ATIS}     & 5,280           & 947             & 1               & 1            & 4           & Crowd-sourced                    \\
\multicolumn{1}{c|}{Scholar}  & 817             & 193             & 1               & 1            & 6           & Crowd-sourced                    \\
\multicolumn{1}{c|}{Academic} & 196             & 185             & 1               & 1            & 6           & Crowd-sourced                    \\
\multicolumn{1}{c|}{GeoQuery} & 877             & 247             & 1               & 1            & 6           & Crowd-sourced                    \\
\multicolumn{1}{c|}{WikiSQL}  & \textbf{80,654} & \textbf{77,840} & \textbf{26,521} & -            & 3           & Crowd-sourced                    \\
\multicolumn{1}{c|}{Spider}   & 10,181          & 5,693           & 200             & \textbf{138} & 9           & 11 annotators                    \\ \midrule
\multicolumn{1}{c|}{SIGMA}    & 6,000           & 4,180           & 160             & 107          & \textbf{44} & \textbf{12 \textless annotators} \\ \midrule
\multicolumn{1}{l}{} & \multicolumn{1}{l}{} & \multicolumn{1}{l}{} & \multicolumn{1}{l}{} & \multicolumn{1}{l}{} & \multicolumn{1}{l}{} & \multicolumn{1}{l}{}  
\end{tabular}%
}
\end{table}
\vspace{-1.2cm}

\subsection{Built-in Python Executor}
The built-in executor is written in the Python language. It includes an implementation of all 44 patterns in our dataset including Distribution, Plot, Numeric, and Query types. By using the executor, users will be able to extract information from the tables and columns in a similar manner as the SQL programming queries the database. The SQL query ``SELECT age FROM Student'' is equivalent to the Python code labels, where main-kind is query, sub-kinds is select, table is Student, column is age, and extra-param is empty. We have created a web page to execute all the Python code labels at research.mehaimeed.com.

\section{Task Definition}
\vspace{-0.3cm}

Given the user questions and schema tables, we need to generate a Python code snippet that will be inserted later into a Python function in our built-in executor. In order to make this dataset more realistic and more representative of databases in the real world, we define the following three rules: Cross-domain, Structure Match, and Synonyms Awareness. The first 
two challenges are similar to those presented in the Spider \cite{yu2018spider} paper. 

\vspace{-0.3cm}
\subsection{Cross-Domain.}

Most of the previous semantic parsing datasets have been based on single datasets from one domain, such as Academic \cite{li2014constructing} and ATIS \cite{price1990evaluation}. The reason for the models performing well may be that they remember some repetitive words in the questions or because they are overfitting the data, which can adversely affect their performance when being applied to other domains. For this reason, we ensure that the training and testing datasets have different databases. Therefore, models will make correct predictions based on comprehending the semantic context of questions.

\subsection{Structure Match} 
In our Text-to-Code task, models are not required to predict the five Python code components correctly with their values. Because in real-world scenarios, people may be aware of the values that they are looking for, but they do not know how to write a programming language code in order to retrieve the data from a database. Additionally, due to the complex nature of the values in some databases or knowledge bases, a prior understanding of the domain is required. There are many models that are good at predicting the structure of the samples. However, they are wrong about the values inside those samples. Therefore, we have included the structure match rule, where the models are required to predict only the five Python code components without values.

\subsection{Synonyms Awareness}
 Some models may give accurate prediction results when being trained on a large dataset, because they can recognize repetitive tokens between inputs and outputs. The problem with only remembering tokens' value and position is that it may lead to overfitting to the data, which causes the models to perform well only on that dataset. Hence, we should take into account the context and meaning of the sentence. In order to address this issue, 20\% of all dataset questions have many synonyms words, which help us indicate whether the model understands the meaning of the words or not. Additionally, in the testing dataset, 10\% of the questions have synonyms words for all patterns, tables, and columns, allowing us to determine which model performs better.
 
\subsection{Multi-Patterns} The model should predict the sub-kinds (patterns) in the second Python code components. Our dataset has 40 statistical patterns. For the distribution and plot statistical questions, the model is only required to figure out one pattern. However, in real-world situations, users may request multiple statistical patterns at a time, such as ``Calculate the most common value and median of the student ages.'' This question is answered using the mode and median statistical patterns. Additionally, predicting one pattern for each question is not that difficult for the models. Therefore, in order to increase the difficulty of the task, we define the multi-patterns rule. For the statistical numeric type, the question can be asked for up to three patterns at the same time. This will increase the difficulty of this task and allow us to measure the robustness of the model in distinguishing between 21 statistical numeric patterns.

\section{Evaluation Metrics}

We have used multiple metrics to measure the models performance on this dataset: execution accuracy, structure matching, and synonyms accuracy. For the execution accuracy, the majority of previous research in the semantic parsing field, such as research in the text-to-SQL \cite{dahl1994expanding} \cite{iyer2017learning} \cite{li2014constructing} \cite{price1990evaluation} \cite{yu2019cosql} \cite{yu2018spider}  \cite{yu2019sparc}  \cite{zhong2017seq2sql} and the question-answering tasks \cite{berant2013semantic}  \cite{joshi2017triviaqa} \cite{rajpurkar2018know} \cite{welbl2018constructing} , use this evaluation metric, which requires the prediction of samples with values. In the SIGMA dataset, only 1444 questions require the prediction of 5 Python code component labels along with the values. Those questions are only related to four different patterns, \ie, where, percentile, trimmed mean, and confidence intervals.

Structure matching measures the percentage of model predictions that exactly match the ground truth Python code component labels. However, there are some values in the fifth extra-param label not included in this metric, which are related to the following four patterns, where, percentile, trimmed mean, and confidence intervals. We do not include them because their values vary depending on the users' inputs. For instance, if the question is ``find the 70\% percentile of doctors wages'', the value of 70\% may vary from question to question, based on the users needs. 

Synonyms accuracy measures the robustness of the model in terms of understanding the meaning of the question. In the testing dataset, 10\% of the questions have synonym words for their patterns, tables, and columns. Suppose that the patterns is standard normal, the table is shop, and column is product price. In this case, the question may be phrased as ``show me the z-distribution for the costs of all products in the store''. \\

\section{Methods}

This section shows experiments with three models that were evaluated on our Text-to-Code task. Questions and schemas will be the inputs to the models, which will then predict the five Python code components. We test our dataset on three models (LGESQL \cite{cao2021lgesql}, SLSQL \cite{lei2020re}, and SmBoP \cite{rubin2020smbop}), which are popular models in the Spider \cite{yu2018spider} dataset competition.

\paragraph{\bf LGESQL} \cite{cao2021lgesql} consists of three parts, \ie, input, hidden, and output modules. An initial embedding of the graph nodes and edges is provided by the input module. The representation of these embeddings is obtained using either word embedding Glove \cite{pennington2014glove} or pre-training language models such as BERT \cite{devlin2018bert} and ELECTRA \cite{clark2020electra}. The second (hidden) module uses graph neural networks to encode and capture the relational structure between the initially generated node embeddings. In the last and output module, a grammar-based syntactic neural decoder is applied, which will construct the abstract syntax tree (AST) of the predicted query. Our modification is primarily focused on generating the desired AST. All 44 patterns have been included in the grammar rules. The major changes involve converting the five Python code components into an AST in the model inputs, then unparsing the AST into five Python code labels in the model outputs.

\paragraph{\bf SLSQL.}  We use a simple base model of SLSQL \cite{lei2020re}, which is composed of two parts: encoder and decoder. The encoder will concatenate the input question and schema items, passing them to the BERT transformer to generate word embeddings. After that, in a two-step decoding process, the decoder first constructs the query without aggregation function and then uses the GRU network to get the final query with aggregation function. As for our dataset, the main modification was made to the input sequence generated by the SLSQL during the preprocessing phase. The input sequence represents the five Python code labels that correspond to each question during the training phase. Furthermore, the SLSQ implemented some constraints during the decoding process, which will ensure that the generated SQL query can be executed. We have also created constraints, in the same manner, to ensure that the five Python code labels obtained can be executed. 

\paragraph{\bf SmBoP} and \textbf{T5.} SmBop \cite{rubin2020smbop} model uses a bottom-up method to solve this problem by constructing the top-K sub-trees of depth $\le$ t at decoding step t during beam search phase. At step t+1, new trees with more depth are constructed from the top-K sub-trees at step t and only K best results will be saved for step t+1 as well. The bottom-up method helps solve an issue from top-down method that before finishing constructing the whole tree, a partial tree cannot provide clear semantic meaning. For the encoder part, SmBop model inherits from RATSQL+GRAPPA \cite{wang2019rat} \cite{yu2020grappa} encoder. Similar as other models, SmBop uses grammar rules to generate results. However, the results we get by using new grammar rules for SmBop are not good. To fully utilize the ability of SmBop model, we use WHERE clauses for sub-kinds to predict instead of using new rules. To help further increase the accuracy of the results, we treat the problem as a text summarization problem. A T5 \cite{raffel2020exploring} model is trained in the summary mode with the natural language question as the text and concatenation of sub-kind names and extra-param as the summary. Then the results from T5 model area used to replace partial results from the previous model.

\section{Experimental Results}
 
% \usepackage{booktabs}
% \usepackage{graphicx}
\begin{table}[]
\centering
\caption{The Structure and Execution Accuracy on the prediction of five Python code components with different patterns types. ``Statistical'' refers to the accuracy of numeric, distribution, and plot types combined. ``Synonyms'' refers to the testing questions with synonyms words for the table, columns and patterns types.}
\label{tab:results}
\resizebox{\textwidth}{!}{%
\begin{tabular}{@{}cccccccc@{}}
\toprule
\multicolumn{8}{c}{\textbf{Structure Accuracy}}                                                                                                                 \\ \midrule
\multicolumn{1}{c|}{\textbf{}}           & \multicolumn{6}{c|}{\textbf{Categorical Results}}                                                         &              \\ \cmidrule(r){1-7}
\multicolumn{1}{c|}{\textbf{Models}}     & Numeric          & Distribution     & Plot    & Query   & Statistical & \multicolumn{1}{c|}{Synonyms} & \textbf{All} \\ \cmidrule(l){2-8} 
\multicolumn{1}{c|}{SLSQL}               & 53.33\%          & 60.56\%          & 59.68\% & 41.81\% & 55.58\%     & \multicolumn{1}{c|}{33.75\%}  & 48.75\%      \\
\multicolumn{1}{c|}{SmBoP + GraPPa + T5} & 63.33\%          & 69.01\%          & 79.03\% & 87.40\% & 66.75\%     & \multicolumn{1}{c|}{37.50\%}  & 77.00\%      \\
\multicolumn{1}{c|}{LGESQL + Glove}      & 67.28\%          & 71.83\%          & 65.08\% & 71.36\% & 67.74\%     & \multicolumn{1}{c|}{42.50\%}  & 69.75\%      \\
\multicolumn{1}{c|}{LGESQL + BERT}       & \textbf{78.00\%} & \textbf{85.92\%} & 80.95\% & 84.67\% & 79.85\%     & \multicolumn{1}{c|}{65.00\%}  & 82.50\%      \\
\multicolumn{1}{c|}{LGESQL + ELECTRA} &
  \textbf{78.00\%} &
  \textbf{85.92\%} &
  \textbf{82.54\%} &
  \textbf{86.18\%} &
  \textbf{80.09\%} &
  \multicolumn{1}{c|}{\textbf{73.75\%}} &
  \textbf{83.37\%} \\ \midrule
\multicolumn{8}{c}{}                                                                                                                                            \\
\multicolumn{8}{c}{\textbf{Execution Accuracy}}                                                                                                                 \\ \midrule
\multicolumn{1}{c|}{SmBoP + GraPPa + T5} & 63.33\%          & 69.01\%          & 79.03\% & 86.14\% & 66.75\%     & \multicolumn{1}{c|}{37.50\%}  & 76.38\%      \\ \midrule
\multicolumn{1}{l}{} &
  \multicolumn{1}{l}{} &
  \multicolumn{1}{l}{} &
  \multicolumn{1}{l}{} &
  \multicolumn{1}{l}{} &
  \multicolumn{1}{l}{} &
  \multicolumn{1}{l}{} &
  \multicolumn{1}{l}{} \\ \bottomrule
\end{tabular}%
}
\end{table}

%The first two evaluation metrics were similar to those used in the Spider \cite{yu2018spider} competition.

The structure accuracy and execution accuracy were used to evaluate the model’s accuracy between ground truth and predicted Python labels.  A comparison of the performance of the three models can be found in \autoref{tab:results}. The SLSQL performs the lowest at 48.75\%, due to the fact that the SLSQL base model does not well solve the issue of schema linking. The accuracy of LGESQL with word vectors Glove \cite{pennington2014glove}, LGESQL with PLM BERT \cite{devlin2018bert}, and LGESQL \cite{cao2021lgesql} with PLM ELECTRA \cite{clark2020electra} is 69.75\%, 82.50\%, and 83.37\%, respectively. There is a problem with Glove in that it does not take into account the context of each word, which explains its lower performance. The PLM ELECTRA outperforms the PLM BERT. This is due to BERT’s replacement of some input tokens with [MASK], which has prevented the model from learning from all inputs. In contrast to BERT, ELECTRA uses Replaced Token Detection to learn from all inputs and determine if each word in the input is substituted, based on its context. 

Since the SmBoP model \cite{rubin2020smbop} relies on the PLM GraPPa \cite{yu2020grappa}, which has been pre-trained on synthetic question-SQL pairs, the results were not satisfactory. However, this issue has been resolved after we use the T5 model \cite{raffel2020exploring} to predict the sub-kinds and extra-params. The model achieves 77\% for structure matching and 76.38\% for execution accuracy. The results of execution accuracy indicate that it performs well, but there are still improvement space to be made. Despite the fact that LGESQL + ELECTRA achieve the highest accuracy in structure match, the percentage of synonyms' accuracy is 73.75\%. As a result, it is evident that there is a need to improve the model in order to handle synonyms words. It is important to note that the prediction of the five Python labels codes with values requires that the model has a prior knowledge of the domain. However, we are more concerned about structure matching, which does not require the prediction of Python labels with values, that is why we have used 3 models for structure matching and one model for execution accuracy.

\section{Conclusions}

We introduce SIGMA, a dataset for Text-to-Code semantic parsing with statistical analysis, which contains 6,000 questions with corresponding Python code over 160 databases. In SIGMA, 44 patterns are covered in our dataset, where 40 of them are used for statistical analysis. With our built-in Python executor, a user can execute the generated Python code. There are three models used in the experiment: LGESQL, SLSQL, and SmBoP. T5 model is used to help improve the accuracy for sub-kinds and extra-params in the SmBoP model. To the best of our knowledge, this is the first dataset for the Text-to-Code task that utilizes Python programming language to retrieve information and perform statistical analysis.
\footnotetext{Citation: S. Almohaimeed, S. Liu, M. Alsofyani, S. Almohaimeed and L. Wang, "SIGMA: A Dataset for Text-to-Code Semantic Parsing with Statistical Analysis," 2023 International Conference on Machine Learning and Applications (ICMLA), Jacksonville, FL, USA, 2023, pp. 851-857, doi: 10.1109/ICMLA58977.2023.00125. IEEE Xplore link: https://ieeexplore.ieee.org/abstract/document/10459845}
\bibliographystyle{splncs04}
\bibliography{sample}

\end{document}